\documentclass[letterpaper, 10 pt, conference]{ieeeconf}  

\IEEEoverridecommandlockouts                              

\usepackage{graphics} 
\usepackage{epsfig} 
\usepackage{mathptmx} 
\usepackage{times} 
\usepackage{amsmath} 
\usepackage{amssymb}  

\usepackage{hyperref}       
\usepackage{url}            
\usepackage{booktabs}       
\usepackage{amsfonts}       
\usepackage{nicefrac}       
\usepackage{microtype}      
\usepackage{lipsum}
\usepackage{tikz}
\usepackage{amsmath}
\usepackage{graphics}
\usepackage{subcaption}
\usepackage{makecell}
\usepackage{hyperref}
\usepackage{xcolor}
\usepackage{cleveref}
\usepackage{pifont}
\newcommand{\xmark}{\ding{55}}%
\usepackage{multirow}
\usepackage{rotating}
\usepackage{pgfplots}
\pgfplotsset{compat=1.8}
\usepackage{eurosym}      
\usepackage{filecontents} 
\usepackage{tikz}
\def\checkmark{\tikz\fill[scale=0.4](0,.35) -- (.25,0) -- (1,.7) -- (.25,.15) -- cycle;}

\newcommand\crule[3][black]{\textcolor{#1}{\rule{#2}{#3}}}
\definecolor{road}{rgb}{1.0, 0.08, 0.58}
\definecolor{side-walk}{rgb}{0.29, 0.0, 0.51}
\definecolor{parking}{rgb}{0.8, 0.6, 0.8}
\definecolor{car}{rgb}{0.0, 0.4, 0.65}
\definecolor{bicyclist}{rgb}{0.63,0.17,0.53}
\definecolor{pole}{rgb}{0.82, 0.78, 0.50}
\definecolor{vegetation}{rgb}{0.13, 0.55, 0.13}
\definecolor{terrain}{rgb}{0.55,0.81,0.35}
\definecolor{trunk}{rgb}{0.64,0.28,0.0}
\definecolor{building}{rgb}{0.83,0.65,0.0}
\definecolor{other-structure}{rgb}{0.82,0.48,0.0}
\definecolor{other-object}{rgb}{0.17,0.87,0.87}

\title{\LARGE \bf
Lite-HDSeg: LiDAR Semantic Segmentation Using Lite Harmonic Dense Convolutions}

\author{
  Ryan Razani*, Ran Cheng*, Ehsan Taghavi, and Liu Bingbing\\
  Huawei Noah's Ark Lab, Toronto, Canada\\
  \texttt{\{ryan.razani, ran.cheng1, ehsan.taghavi, liu.bingbing\}@huawei.com} 
  \thanks{$^*$ indicates equal contribution.}
  }

\begin{document}

\maketitle
\thispagestyle{empty}
\pagestyle{empty}


\begin{abstract}

Autonomous driving vehicles and robotic systems rely on accurate perception of their surroundings. Scene understanding is one of the crucial components of perception modules. Among all available sensors, LiDARs are one of the essential sensing modalities of autonomous driving systems due to their active sensing nature with high resolution of sensor readings. Accurate and fast semantic segmentation methods are needed to fully utilize LiDAR sensors for scene understanding. In this paper, we present Lite-HDSeg, a novel real-time convolutional neural network for semantic segmentation of full $3$D LiDAR point clouds. Lite-HDSeg can achieve the best accuracy vs. computational complexity trade-off in SemanticKitti benchmark 
and is designed on the basis of a new encoder-decoder architecture with light-weight harmonic dense convolutions as its core. Moreover, we introduce ICM, an improved global contextual module to capture multi-scale contextual features, and MCSPN, a multi-class Spatial Propagation Network to further refine the semantic boundaries. 
Our experimental results show that the proposed method outperforms state-of-the-art semantic segmentation approaches which can run real-time, thus is suitable for robotic and autonomous driving applications.

\end{abstract}


\section{Introduction}
Perception is one of the primary tasks for autonomous driving. In perception for autonomous vehicles, one main challenge can be generalized as object recognition on road. Examples of such objects include but not limited to cars, cyclists, pedestrians, traffic signs, derivable area and sidewalks etc. Due to lack of bounding box abstraction for many of these classes and the need to recognize all these objects in one unified end-to-end approach, a semantic segmentation approach applies, which attempts to predict the class label or tag, for each pixel of an image or point of a point cloud.

To achieve this task, many sensors can be leveraged in robotic systems, such
as cameras (mono, stereo), Light Detection And Ranging (LiDAR), ultrasonic, and
Radio Detection And Ranging (RADAR) sensors. Among these sensors, LiDAR becomes the sole choice of sensor for perception tasks such as $3$D semantic segmentation because of its active sensing nature with high resolution of $3$D sensor readings. 

\begin{figure}[htb]
    \centering
    \includegraphics[width=0.45\textwidth]{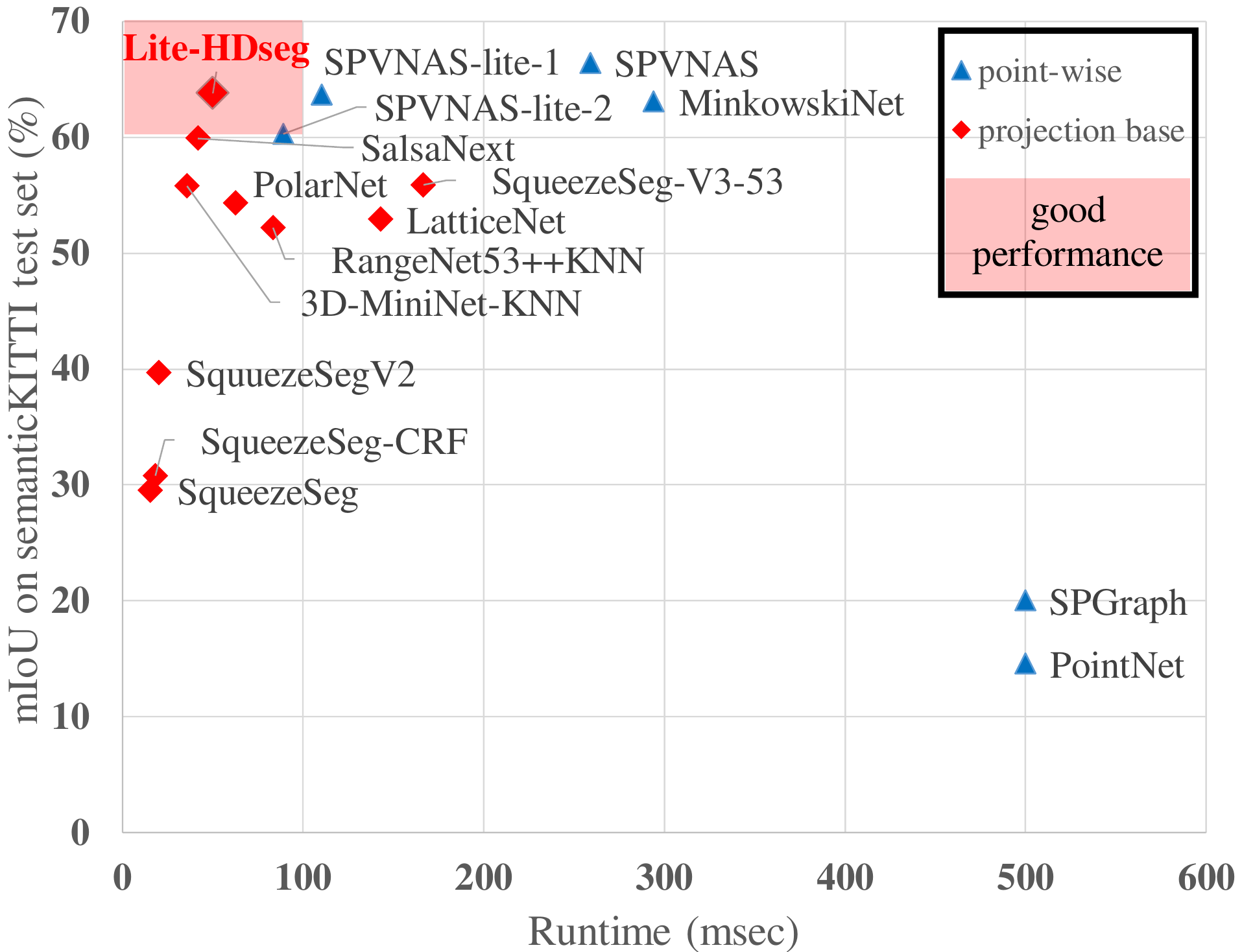}
    \caption[Iou-vs-Speed]{Accuracy (mIoU) vs. runtime. Only methods with less than $500$ms runtime are shown for better representation.}
    \label{Iou_vs_Speed}
    \vspace{-10px}
\end{figure}

During the past few years, and with availability of recent new datasets such as SemanticKITTI \cite{DBLP:conf/iccv/BehleyGMQBSG19}, many new methods have been proposed to address the problem of LiDAR semantic segmentation. Although different in design and solution proposal, many of the methods can be categorized into these four groups: point-based methods such as \cite{thomas2019kpconv}, image-based methods such as \cite{iandola2016squeezenet}, voxel-based methods such as \cite{tchapmi2017segcloud} and graph-based methods such as \cite{landrieu2018large}. Section \ref{related_work} is dedicated to covering different methods in literature on the subject of LiDAR semantic segmentation.

Overall, point-based methods can achieve better accuracy at the cost of high computational complexity and memory consumption. A more reasonable and realistic choice is image-based or voxel-based methods where raw and unordered point cloud is projected onto a structured plane. Lite-HDSeg is an end-to-end LiDAR semantic segmentation method that is solely based on range-image (a transformation of $3$D LiDAR point clouds into an image using spherical projection). Thereby, it benefits from minimal prepossessing and efficient implementation using standard CNNs. The contributions of this paper are summarized as follows:

\begin{itemize}
  \item A novel residual Inception-like Context Module (ICM) to capture global and local context information in full $360$ degrees LiDAR scan.  Incorporating ICM at the beginning of the neural network improves learning local and global features for the task of semantic segmentation as it is shown in Section \ref{sec:ablation};
  \item A new and improved encoder-decoder CNN model on a new encoder named Lite-HDSeg, a residual network decoder, a multi-class SPN and a modified boundary loss which is trained end-to-end using the spherical data representation (range-image). Lite-HDSeg surpasses state-of-the-art methods in accuracy vs. runtime trade-off on a large-scale public benchmark, SemanticKITTI (shown in Fig. \ref{Iou_vs_Speed});  
    \item A thorough analysis on the LiDAR semantic segmentation performance based on different ablation studies, and  qualitative and quantitative results. 
\end{itemize}

The rest of the paper is organized as follows. In Section \ref{related_work}, a brief history of recent and related works is given. Section \ref{sec:proposed_method} presents the proposed Lite-HDSeg network with detailed description of its components and novel techniques used to design the model. Section \ref{sec:result} demonstrates a comprehensive experimental results and thorough qualitative and quantitative evaluation on the public benchmark, SemanticKITTI \cite{DBLP:conf/iccv/BehleyGMQBSG19}. Finally, some conclusions are drawn in Section \ref{sec:conclusion}.
\section{Related Work}
\label{related_work}

\textbf{Point-based methods.} 
A natural approach to extract features from point cloud data is to treat it in its original form, i.e., unordered point cloud, directly process it using various approaches such as a neural network. In this way, all data points will be processed and a label is predicted for each input data point. 
The pioneering methods of this group are PointNet \cite{qi2017pointnet} and PointNet++ \cite{qi2017pointnet++} which use shared Multi-Layer Perceptrons (Shared MLPs) to extract features for different tasks such as semantic segmentation.

Much research focuses to efficiently implement point-based methods without sacrificing performance \cite{hu2019randla}. However, point-based approaches remain successful in tasks with only small input point clouds \cite{hu2019randla}. KPConv \cite{thomas2019kpconv} develops a flexible way to use any number of kernel points and thus can be extended to deformable convolutions that can use these flexible kernel points to learn local geometry. More recently, KPRNet \cite{kochanov2020kprnet} has combined ResNext \cite{xie2017aggregated} and \cite{thomas2019kpconv} to achieve better results, however, the computational complexity of the model makes it less attractive for autonomous driving/robotic applications.

\textbf{Image-based methods.} 
\label{spherical approaches}
To benefit from $2$D convolutional neural networks such as \cite{DBLP:journals/corr/RonnebergerFB15}, one can project $3$D LiDAR data onto a $2$D surface using spherical projection \cite{iandola2016squeezenet}. Using spherical projection turns the point cloud segmentation task into image segmentation problem, in which a prediction can be made for each of the projected data points.

Image-based methods, in specific those which use spherical projection, perform better in terms of segmentation accuracy vs computational complexity trade-off. This is due to leveraging $2$D convolutional neural networks (CNNs), such as residual blocks \cite{he2016deep}, depth-separable convolutions \cite{chollet2017xception}, and dilated convolutions \cite{yu2015multi}. Therefore, spherical projection methods have gained attention in recent years and been used in many methods such as SqueezeSeg series \cite{wu2018squeezeseg, wu2019squeezesegv2, xu2020squeezesegv3}, RangeNet++ \cite{milioto2019rangenet++}, DeepTemporalSeg \cite{dewan2019deeptemporalseg}, SalsaNet \cite{aksoy2019salsanet}, SalsaNext \cite{cortinhal2020salsanext}, etc. SalsaNext \cite{cortinhal2020salsanext} inherits the encoder-decoder architecture from SalsaNet \cite{aksoy2019salsanet}, which contains a series of residual blocks \cite{he2016deep} in the encoder and upsamples features in the decoder part. An extra layer, added before the encoder, brings in more global context information, together with other techniques such as average pooling and dropouts. A new loss term, Lov\'asz-Softmax loss \cite{berman2018lovasz}, is also used in SalsaNext \cite{cortinhal2020salsanext} which can optimize the mean Intersection over Union (IoU) metric directly. With all the above-mentioned changes, SalsaNext achieves a higher score on SemanticKITTI leaderboard than other publicly available methods in this category.

\begin{figure*}[ht!]
    \centering
    \includegraphics[trim = 0in 0in -1in 0in, clip, width=5.7in]{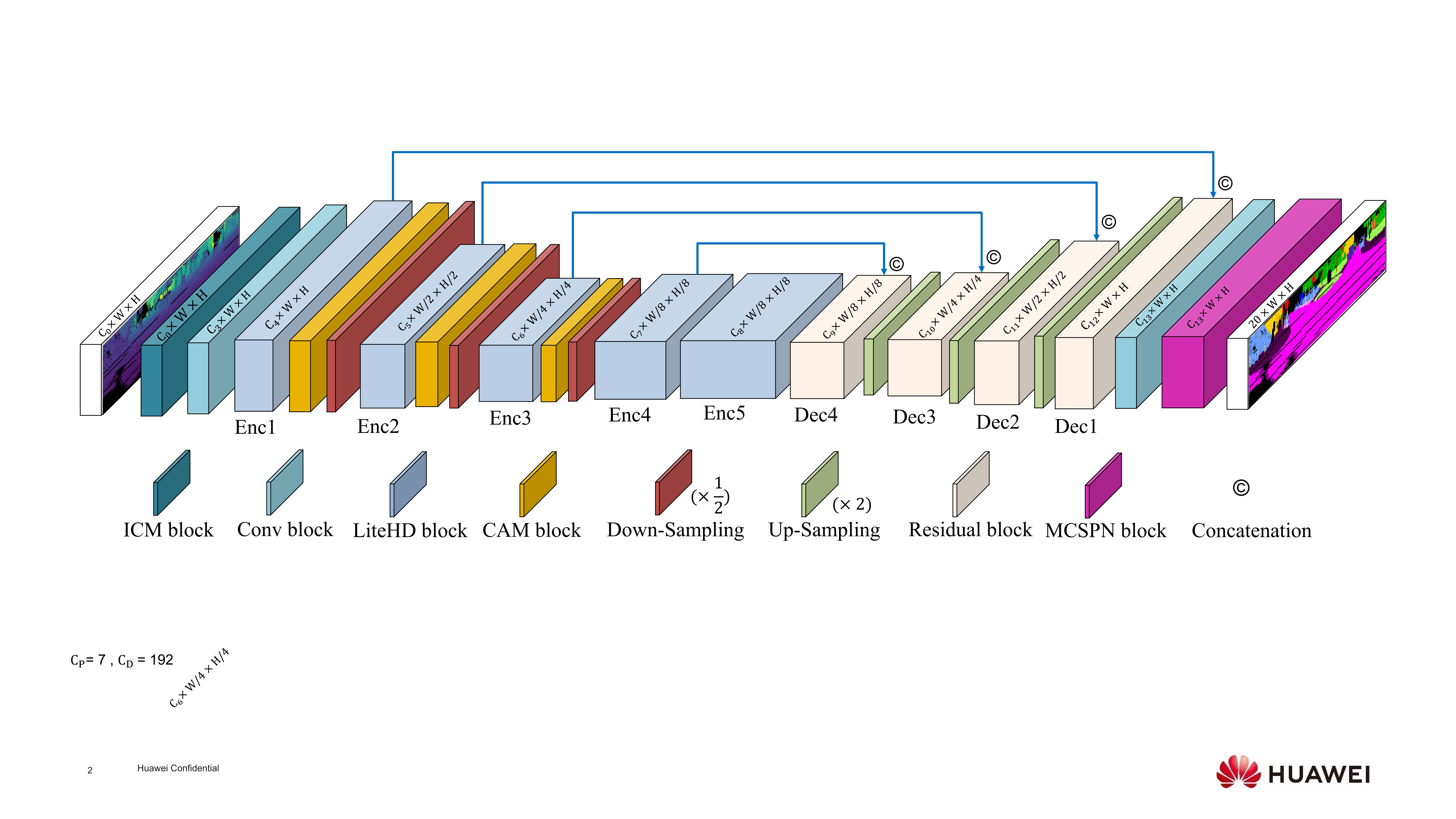}
    \caption[Lite-HDSeg]{LiteHDSeg Architecture.}
    \label{Lite-HDSeg}
    \vspace{-8px}
\end{figure*}

\textbf{Voxel-based methods.} Voxel-based methods use $3$D data points and project them into cubical voxels of known size. Each voxel, which may contain more than one data point, carries different information, such as maximum height, mean, variance, and other information related to the data points associated with each voxel. Voxel-based methods usually take voxels as input and predict each voxel with one semantic label. SEGCloud \cite{tchapmi2017segcloud} is one of the early data driven methods to use voxel representation for point cloud semantic segmentation. 
ScanComplete \cite{Dai_2018_CVPR} takes in a $3$D scan of a scene and outputs a complete $3$D model with a semantic label for every voxel with a specially designed $3$D CNN model, whose filter kernels are invariant to the overall scene size. More recently, the Minkowski CNN \cite{Choy_2019} has been proposed which can efficiently handle $3$D sparse convolutions and other standard neural network layers and thus reduces the computational costs of $3$D CNNs significantly. A neural architecture search (NAS) model based on \cite{Choy_2019} also introduced recently, i.e. SPVNAS \cite{tang2020searching}, that can achieve state-of-the-art accuracy at the cost of high computational cost. Although efficient $3$D CNNs are available, voxel-based methods suffer from loosing granular data and are prone to detecting smaller objects on the scene, specially with larger voxels.

\textbf{Graph-based methods.} A graph representation can be reconstructed from point cloud data, where a vertex represents a point or a group of points, and edges represent adjacency relationship between vertices. 
Using graph representation, authors in \cite{landrieu2018large} introduced a neural network model that benefits from geometrically homogeneous elements from the scanned scene using a structure called superpoint graph (SPG). More recently, GACNet \cite{wang2019graph} proposed a graph convolution with new kernels that can be shaped dynamically to adapt to the structure of an object. In \cite{jiang2019hierarchical}, authors designed a neural network  by proposing a technique to extract features, benefiting from a hierarchical graph framework to extract point/edge level features. Although theoretically sound, graph-based methods are not suitable for real-time robotic applications due to their complex nature, and more research needs to be done to make them a practical choice.

\vspace{-0.3mm}

\section{Proposed Method} 
\label{sec:proposed_method}

To achieve real-time performance and benefit from conventional convolution functions commonly used in computer vision, we use spherical projection to project the LiDAR point cloud into a spherical coordinate system, creating a $2$D range-image. Consequently, the point cloud segmentation task can be approached as an image segmentation problem. To accomplish this, one can use the following function to map all $3$D points into a bitmap image ($\mathbb{R}^3 \rightarrow \mathbb{R}^2$) as,

\begin{equation}
    \label{proj}
 	\left(
 	\begin{matrix}
 	u \\
 	v
 	\end{matrix}
 	\right) =
 	\left(
 	\begin{matrix}
 	\frac{1}{2}[1-arctan(y,x)\pi^{-1}]W \\
 	[1-(arcsin(zr^{-1})+f_{up})\frac{1}{f}]H
 	\end{matrix}
 	\right)
\end{equation}

\noindent where $W$ and $H$ are the width and height of the range-image. The vertical field of view of the sensor is $f = f_{up} + f_{down}$. Symbols $u$ and $v$ are the horizontal and vertical coordinates of the corresponding pixel in the bitmap image, respectively.
 
Using \eqref{proj}, one can generate an input of arbitrary size $W$ and $H$ with channels $(x,y,z,rem,r)$, where $(x,y,z)$ are the Cartesian coordinates of a data point, $rem$ is the remission or intensity reading from the sensor and $r$ is the range, respectively. In the following subsections, a novel neural networks architecture and its design is explained to address this problem along with experimental results and ablation studies.

The proposed network structure is based on a) multi-scale convolutional learning module, Inception-like Context Module (ICM), to extract multi-scale contextual  features  from  range-image  to  be  processed  by the  encoder-decoder  network, b) lite version of HarDNet \cite{chao2019hardnet} as the encoder with less dense connections to achieve better performance (see Fig.~\ref{fig:LHD}), c) CAM module in the encoder to collect nearby context from neighborhood feature maps, d) Multi-class Convolutional Spatial Propagation Network (MCSPN) before the last layer of convolutions in the encoder-decoder CNN to refine the masks of each class for the segmentation task and e) a dedicated boundary loss in addition to cross entropy and Lov\'asz loss loss for the segmentation task of LiDAR point cloud data.
The combination of all above-mentioned contributions demonstrate that segmentation results can substantially be improved over the range map with a large margin. 

The general block diagram of the proposed model is shown in Fig.~\ref{Lite-HDSeg} and the details of the network architecture are described in subsections below.

\subsection{Context Feature Extractor}
\begin{figure} [!htbp]
	\centering
	\begin{subfigure}[t]{0.45\textwidth}
		\includegraphics[width=\textwidth]{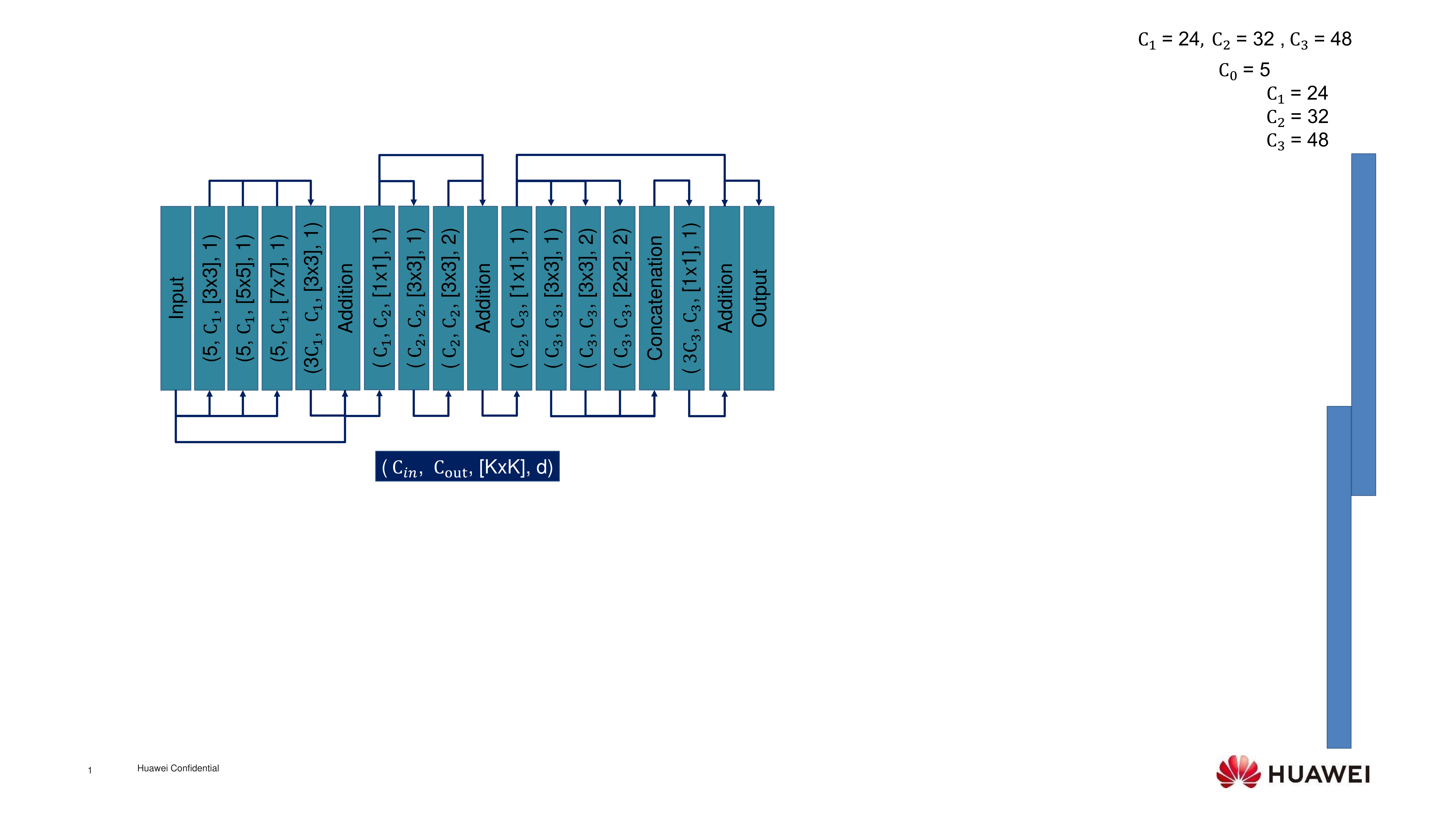}
		\caption{ICM block}
		\label{fig:ICM}
	\end{subfigure}
	~
	\begin{subfigure}[t]{0.26\textwidth}
		\includegraphics[width=\textwidth]{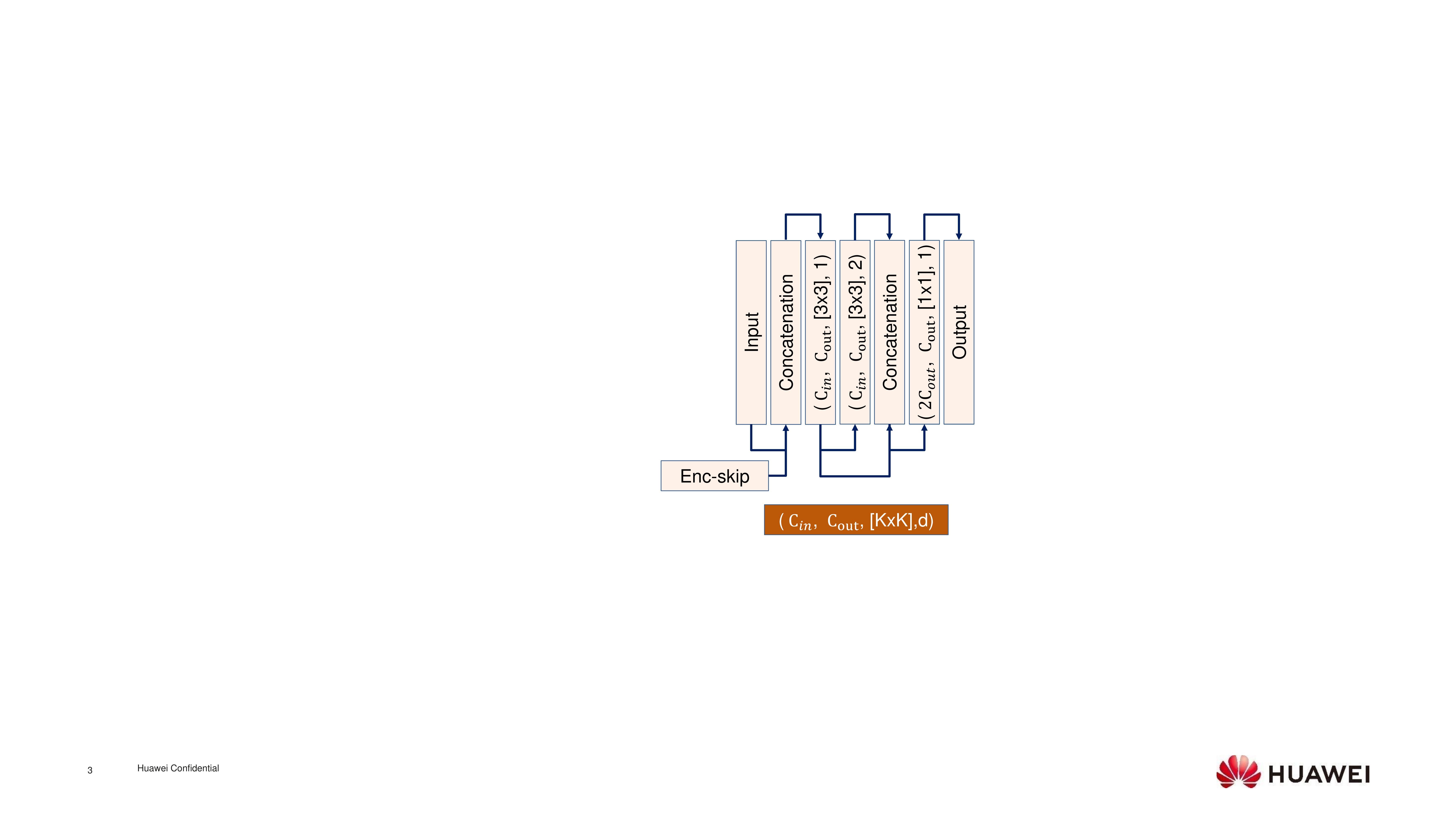}
		\caption{Residual block}
		\label{fig:Decoder_Res}
	\end{subfigure}
	~ 
	\caption[ICM]{(a) Inception-like Context Module (ICM). (b) Residual Block.  Parameters of the $2$D convolution layers are input channels, output channels, kernel size and dilation ratio for $C_{in}$, $C_{out}$, $[K\times K]$ and $d$, respectively. The input to the ICM is the range-image with features $(x,y,z,rem,r)$ with the size $[C_{0}, W, H]$. The output of the ICM has $48$ output channels with spatial size $[W,H]$. }
    \vspace{-10px}
\end{figure}

We present a multi-scale convolutional learning module, Inception-like Context Module (ICM), to extract multi-scale contextual features from range-image to be processed by the encoder-decoder network targeted at semantic segmentation. The proposed context feature extractor consists of concatenation of several multi-scale convolution layers with residual connections and larger receptive fields to extract rich global information which is essential in learning complex correlations between classes with different size. This yields to gather the spatial fine-grained information along with global context information. A detailed visual description with numerical values for channels, kernels and dilation ratios can be found in Fig.~\ref{fig:ICM}.

\subsection{Lite-HDSeg Encoder}
\label{encoder}
\begin{figure} [!htbp]
	\centering
	\begin{subfigure}[t]{0.4\textwidth}
		\includegraphics[width=\textwidth]{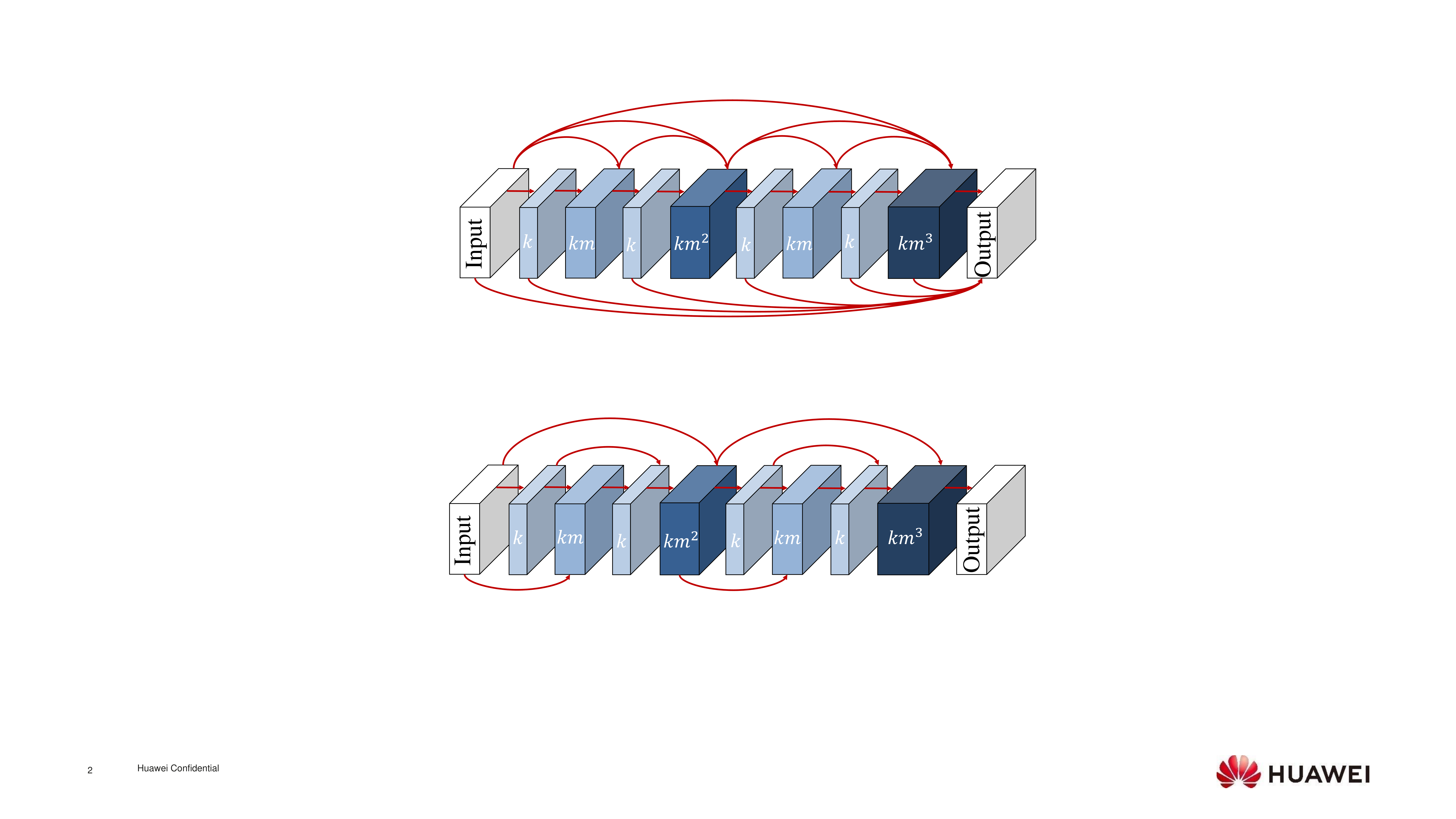}
		\caption{Original HD block}
		\label{fig:HD}
	\end{subfigure}
	~
	\begin{subfigure}[t]{0.4\textwidth}
		\includegraphics[width=\textwidth]{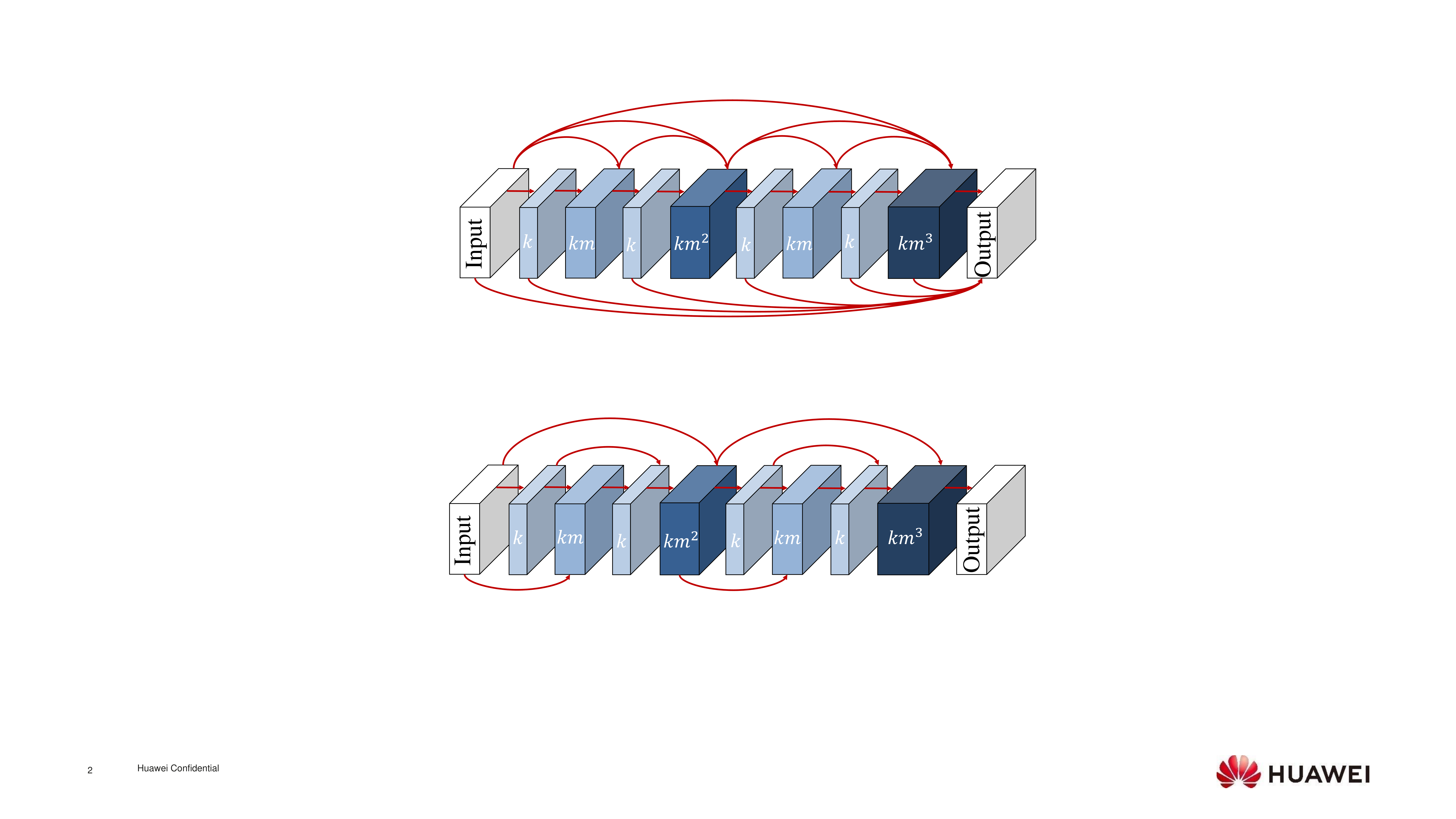}
		\caption{ Proposed Lite-HD block}
		\label{fig:LHD}
	\end{subfigure}
	~ 
	\caption[HarDnet]{Comparison of a) original HD block and (b) proposed Lite-HD block}
	\vspace{-10px}
\end{figure}

Using shortcuts or skip connections proved to be an effective design choice in tasks such as semantic segmentation. Shortcuts enable implicit supervision to make networks deepen without degradation. DenseNets \cite{huang2017densely} proposed a shortcut scheme where all preceding layers are concatenated together. Authors in \cite{huang2017densely} showed that this novel shortcuts scheme improves the deep supervision and achieves good results in image segmentation task. log-DeseNets \cite{hu2017log} and SparseNet \cite{liu2018sparsenet} both attempt to sparsify the DenseNets by sparsely connecting layers together. Although smaller set of shortcuts can achieve fast convergence speed, they both need to increase the growth rate (output channel width) to recover from connection pruning. One of the most successful design in this domain is Harmonic Dense Net (HDNet) \cite{chao2019hardnet}. HarDNet proposed shortcuts following power-of-two-th harmonic waves achieving a significant reduction in cost, while keeping an acceptable performance of the model.  As shown in Fig.~\ref{fig:HD}, layer $k$ connects to layer $k-2^n$ if $2^n$ divides $k$, where $n$ is a non-negative integer, $k-2^n \geqslant 0$ and the multiplier $m$ serves as a low-dimensional compression factor. Further details can be found in \textsection 3 of \cite{chao2019hardnet}.
Overall, compared to the dense blocks, which connects all the layers iteratively in a block, HD block (Harmonic Dense block) has much deeper structure yet, uses less skip connections (shortcuts) according to the harmonic-pattern to propagate the gradients. HD block was inspired from the log-DenseNet \cite{hu2017log} which cut the network complexity yet, kept the effective skip connectivity that ensured the propagation of gradient.

Our proposed Lite-HDSeg uses a series of simplified HD blocks for the encoder by removing some of the skip connections, while keeping the critical connections (Enc$1$--Enc$5$). This is due to the fact that the dense skip connections introduce many redundant and repeated channels. This systematic pruning in the skip connections reduces the computational complexity and improves memory efficiency, which are the cornerstone of real-time applications. In the proposed Light HD Block (LHD), each convolution layer $f_i$ takes a direct input from at most $\lfloor \frac{log(i)}{5} \rfloor$ number of previous layers, and these input layers are apart from layers with base $5$, i.e.:
\begin{equation}
    x_i = f_{i}(\textrm{concat}({x_{\lfloor i-2 | i-5^{k}+1 \rceil} : k=1,...,\lfloor log(i) \rfloor}); \mathnormal{\theta_i})
\end{equation}
where $\lfloor \cdot \rfloor$ is the floor function, $|$ is \textbf{or} operation, and $\lfloor \cdot \rceil$ is the nearest integer function. For example, the feature maps of layer $i$ is concatenated from layers
\begin{equation*}
\big\{ i-2, \lfloor \frac{log(i)}{5} \rfloor, \lfloor log(\frac{log(i)}{5})/5 \rfloor, ...\big\}
\end{equation*} 
Finally, the overall complexity of a Lite HDBlock is
\begin{equation}
\sum^{L}_{i=1}(\frac{log(i)}{5} + 1) \leq L + \frac{1}{5}L \> logL = \Theta(\frac{1}{5}L \> logL)
\end{equation} 
which is pruned significantly compared to the DenseNet \cite{huang2017densely} and HarDNet \cite{chao2019hardnet}. Our experimental results show that pruning redundant connections in HD block not only increases the model runtime efficiency, but also improves the model performance substantially.

\subsection{Context Aggregation Module}
Context Aggregation Module (CAM) is adopted from the SqueezeSegV2 \cite{wu2019squeezesegv2} and is used after each LHD block as an attention map to emphasize the feature maps locally. As stated in \cite{wu2019squeezesegv2}, the noise that is in the nature of LiDAR point cloud (missing points in the range-image, sensor jitters and mirror reflections) can cause a CNN model to perform poorly. Since the CAM module collects the nearby context from neighborhood feature maps, these characteristics of the sensor can be compensated at the time of training, generating features that are robust against LiDAR specific noise.

\subsection{Decoder}
\label{decoder}
The proposed decoder is based on residual blocks (Res-block). The proposed residual block is a combination of conv-blocks in a specific order and increased kernel size as shown in Fig.~\ref{fig:Decoder_Res} with a skip connection from the encoder with the same spatial size. The Resblock in Fig.~\ref{fig:Decoder_Res} along with an up-sampling layer is used four times (Dec4, Dec3, Dec2 and Dec1) to generate an output with the spatial size of the input range-image. The final layer in the decoder is a convolutional layer to generate output with the depth size  $C_{13}$. 

\subsection{Mutli-class Spatial Propagation Network}
\label{mcspn}
To improve the performance of segmentation result, we adapt convolutional spatial propagation network and extend this framework to support multiple classes. In convolutional propagation network, a convolutional neural network is trained to learn explicit affinity between each pixel and its neighbors (in $2$D image). Then, the segmentation mask is refined according to the learned affinity. Unlike dense CRF \cite{wu2018squeezeseg} in which each pixel is connected with others, the proposed method only contains three way connection, as summarized in the following equation,
\begin{equation}
    h_{k, t, c} = (1- \sum_{k \in \mathit{N}, c \in \mathcal{C}}p_{k, t, c})x_{k,t,c} + \sum_{k \in \mathit{N},\:c \in \mathcal{C}}p_{k, t, c} \> h_{k, t-1, c}
\end{equation}
where $h$ is the propagated hidden feature map, $k$ is current pixel propagation position, $\mathit{N}$ is the neighborhood pixel for one direction, $c \in \mathcal{C}$ is the semantic class, and $t$ is the propagation time step. Parameter $x_{k,t,c}$ is the original output feature map from the decoder, and $p_{k, t, c}$ is the propagated affinity map of one direction (for example, left to right). The final output feature map is then fused from linearly propagated  feature map from last time step through four directions. This propagation refines the semantic of small objects and boosts the overall performance. Finally, the output of the network is passed through a final convolutional layer followed by a $softmax$ function, where the final predictions are being made for each pixel and loss is calculated.
 
\subsection{Loss Function}
To address the specific data type used in this work, the labeled data, and its lack of a balanced ground truth labels for all classes, four different loss terms are used, i.e., weighted cross entropy, Lov\'asz loss, boundary loss,  and weight decay regularization, which are parameterized as ${L}_{wce}$, $L_{ls}$, $L_{bd}$ and $L_{reg}$, respectively. The weighted cross entropy (WCE) loss \cite{zhang2018generalized, panchapagesan2016multi} is suitable where a neural network model deals with multiclass classification problem much like semantic segmentation. The second loss term that is used in this work is Lov\'asz loss \cite{berman2018lovasz}. As shown in \cite{rezatofighi2019generalized, cortinhal2020salsanext}, the Lov\'asz loss is an effective additional loss term that can be used for different machine learning tasks such as object detection and segmentation. 

As stated in \cite{bokhovkin2019boundary}, relying on pixel-level losses alone such as cross-entropy for predicting curvilinear structures is insufficient. In specific to semantic segmentation, there exist complex boundaries between different classes of various shapes which makes cross-entropy and other advanced losses such as Lov\'asz loss sub-optimal for the task. To solve this problem, \cite{bokhovkin2019boundary} proposed to add an extra term to the conventional loss terms, manually extracting boundaries and forcing the network to focus on those regions more than other parts. Therefore, we use boundary loss to account for the error in the boundary of different classes. The loss term  ${L}_{bd}$, used for the task of LiDAR semantic segmentation, can be calculated as,

\begin{equation}
    {L}_{bd}(y,\hat{y})=1 - \frac{2P^cR^c}{P^c + R^c}
\end{equation}
where $P^c$ and $R^c$ are the precision and recall of the predicted boundary map $y^b_{pred}$ with respect to the ground truth boundary map $y^b_{gt}$ for class $c$, respectively. $y$ represents the ground truth and $\hat{y}$ is the model output for each class. The boundary map is thus calculated as,

\begin{equation}
    y^b =pool(1 - y, \theta) - (1-y)
\end{equation}
where $pool(\cdot, \cdot)$ is a pixel-wise max-pooling operation for the binary map within a sliding window, i.e. $\theta$, to extract the vicious boundary. 

Finally, the total loss is computed by combining the previous loss terms with a weight decay regularizer and is defined as,
\begin{align}
    {L}_{tot}(y,\hat{y})=\alpha{L}_{wce}(y,\hat{y})+\beta{L}_{ls}(y,\hat{y})+ \gamma{L}_{bd}(y,\hat{y}) + \lambda{L}_{reg} 
\end{align}
where $\alpha$, $\beta$, $\gamma$, and $\lambda$ are the weights associated to weighted cross entropy, Lov\'asz-Softmax, boundary loss and regularization term, respectively.

\subsection{Post-processing}

To project the prediction labels from range-images back to $3$D representation, a KNN post-processing step is employed.
Every point is given a new label based on its $K$ closest points. However, instead of finding the nearest neighbours in the complete unordered point cloud, a sliding window approach is used to sub-sample the point cloud \cite{milioto2019rangenet++}.

\section{Experimental Results}
\label{sec:result}

To evaluate the performance of the proposed method and to provide experimental results, including ablation studies, a large-scale SemanticKITTI \cite{DBLP:conf/iccv/BehleyGMQBSG19} dataset is used.
SemanticKITTI \cite{DBLP:conf/iccv/BehleyGMQBSG19} is the largest available dataset with fully labeled LiDAR point clouds of various scenes. Due to this, many of the recent works such as \cite{cortinhal2020salsanext,thomas2019kpconv,xu2020squeezesegv3} evaluated their methods based on SemanticKITTI, making it a standard dataset for evaluating LiDAR semantic segmentation methods. Statistical analysis and more information about SemanticKITTI can be found in supplementary documents.

Mean Intersection over Union ($\mathbf{mIoU}$) is used as our evaluation metric to evaluate and compare our methods with others. It is the most popular metric for evaluating semantic point cloud segmentation and can be formalized as 
\begin{equation}
            \label{slice}
         	mIoU = \frac{1}{n}\sum_{c=1}^{n}{\frac{TP_c}{TP_c+FP_c+FN_c}},
\end{equation}
where $TP_c$ is the number of true positive points for class $c$, $FP_c$ is the number of false positives, and $FN_c$ is the number of false negatives. As the name suggests, the IoUs for each class are calculated and then the mean is taken.

\begin{table*}[ht!]
\Huge
\centering
\resizebox{1.75\columnwidth}{!}{
\begin{tabular}{c|l|ccccccccccccccccccccc}

\hline 
Category & 
Method & \begin{sideways} Mean IoU \end{sideways} 
& \begin{sideways} Car \end{sideways} 
& \begin{sideways} Bicycle \end{sideways} 
& \begin{sideways} Motorcycle \end{sideways} 
& \begin{sideways} Truck \end{sideways} 
& \begin{sideways} Other-vehicle \end{sideways} 
& \begin{sideways} Person \end{sideways} 
& \begin{sideways} Bicyclist \end{sideways} 
& \begin{sideways} Motorcyclist \end{sideways} 
& \begin{sideways} Road \end{sideways} 
& \begin{sideways} Parking \end{sideways} 
& \begin{sideways} Sidewalk \end{sideways} 
& \begin{sideways} Other-ground \end{sideways} 
& \begin{sideways} Building \end{sideways} 
& \begin{sideways} Fence \end{sideways} 
& \begin{sideways} Vegetation \end{sideways} 
& \begin{sideways} Trunk \end{sideways} 
& \begin{sideways} Terrain \end{sideways} 
& \begin{sideways} Pole \end{sideways} 
& \begin{sideways} Traffic-sign \end{sideways} 
& \begin{sideways} FPS (Hz) \end{sideways}  \\
\hline
\multirow{5}{*}{\begin{sideways} Point-wise  \end{sideways} }
& LatticeNet  \cite{rosu2019latticenet} 
& $52.9$ & $92.9$ & $16.6$ & $22.2$ & $26.6$ & $21.4$ & $35.6$ & $43.0$ & $\textbf{46.0}$ & $90.0$ & $59.4$ & $74.1$ & $22.0$ & $88.2$ & $58.8$ & $81.7$ & $63.6$ & $63.1$ & $51.9$ & $48.4$  & $7$\\

& Kpconv \cite{thomas2019kpconv} 
& $58.8$ & $96.0$ & $30.2$ & $42.5$ & $33.4$ & $44.3$ & $61.5$ & $61.6$ & $11.8$ & $88.8$ & $61.3$ & $72.7$ & $31.6$ & $90.5$ & $64.2$ & $84.8$ & $69.2$ & $69.1$ & $56.4$ & $47.4$ & $-$ \\

& FusionNet  \cite{zhang12356deep}
& $61.3$  & $95.3$ & $47.5$ & $37.7$ & $\textbf{41.8}$ & $34.5$ & $59.5$ & $56.8$ & $11.9$ & $91.8$ & $68.8$ & $77.1$ & $\textbf{30.8}$ & $\textbf{92.5}$ & $\textbf{69.4}$ & $84.5$ & $69.8$ & $68.5$ & $60.4$ & $\textbf{66.5}$ & $-$\\

& Cylinder3D  \cite{zhou2020cylinder3d}
& $61.8$ & $\textbf{96.1}$ & $\textbf{54.2}$ & $47.6$ & $38.6$ & $\textbf{45.0}$ & $65.1$ & $63.5$ & $13.6$ & $91.2$ & $62.2$ & $75.2$ & $18.7$ & $89.6$ & $61.6$ & $85.4$ & $69.7$ & $69.3$ & $\textbf{62.6}$ & $64.7$ & $-$\\

& KPRNet   \cite{kochanov2020kprnet}
& $63.1$ & $95.5$ & $54.1$ & $\textbf{47.9}$ & $23.6$ & $42.6$ & $\textbf{65.9}$ & $\textbf{65.0}$ & $16.5$ & $\textbf{93.2}$ & $\textbf{73.9}$ & $\textbf{80.6}$ & $30.2$ & $91.7$ & $68.4$ & $\textbf{85.7}$ & $\textbf{69.8}$ & $\textbf{71.2}$ & $58.7$ & $64.1$ & $0.3$\\


\hline
\multirow{8}{*}{\begin{sideways} Projection-based  \end{sideways} }
& DeepTemporalSeg \cite{dewan2019deeptemporalseg} 
& $37.6$ & $81.5$ & $29.4$ & $19.6$ & $6.6$ & $6.5$ & $23.7$ & $20.1$ & $2.4$ & $85.8$ & $8.7$ & $59.3$ & $1.0$ & $78.6$ & $39.6$ & $77.1$ & $46.0$ & $58.1$ & $32.6$ & $39.1$ & $-$ \\

& SqueezeSegV2 \cite{wu2019squeezesegv2} 
& $39.7$ & $81.8$ & $18.5$ & $17.9$ & $13.4$ & $14.0$ & $20.1$ & $25.1$ & $3.9$ & $88.6$ & $45.8$ & $67.6$ & $17.7$ & $73.7$ & $41.1$ & $71.8$ & $35.8$ & $60.2$ & $20.2$ & $36.3$ & $50$\\

& RangeNet53++KNN \cite{milioto2019rangenet++} 
& $52.2$ & $91.4$ & $25.7$ & $34.4$ & $25.7$ & $23.0$ & $38.3$ & $38.8$ & $4.8$ & $\textbf{91.8}$ & $65.0$ & $75.2$ & $27.8$ & $87.4$ & $58.6$ & $80.5$ & $55.1$ & $64.6$ & $47.9$ & $55.9$ & $12$  \\

& PolarNet \cite{zhang2020polarnet} 
& $54.3$ & $93.8$ & $40.3$ & $30.1$ & $22.9$ & $28.5$ & $43.2$ & $40.2$ & $5.6$ & $90.8$ & $61.7$ & $74.4$ & $21.7$ & $90.0$ & $61.3$ & $\textbf{84.0}$ & $65.5$ & $67.8$ & $51.8$ & $57.5$ & $16$ \\

& 3D-MiniNet-KNN \cite{alonso20203d} 
& $55.8$ & $90.5$ & $42.3$ & $42.1$ & $28.5$ & $29.4$ & $47.8$ & $44.1$ & $14.5$ & $91.6$ & $64.2$ & $74.5$ & $25.4$ & $89.4$ & $60.8$ & $82.8$ & $60.8$ & $\textbf{66.7}$ & $48.0$ & $56.6$  & $28$\\

& SqueezeSegV3-53 \cite{xu2020squeezesegv3} 
& $55.9$ & $92.5$ & $38.7$ & $36.5$ & $29.6$ & $33.0$ & $45.6$ & $46.2$ & $20.1$ & $91.7$ & $63.4$ & $74.8$ & $26.4$ & $89.0$ & $59.4$ & $82.0$ & $58.7$ & $65.4$ & $49.6$ & $58.9$ & $6$  \\

& SalsaNext \cite{cortinhal2020salsanext}
& $59.5$ & $91.9$ & $\textbf{48.3}$ & $38.6$ & $\textbf{38.9}$ & $31.9$ & $\textbf{60.2}$ & $59.0$ & $19.4$ & $91.7$ & $63.7$ & $75.8$ & $29.1$ & $90.2$ & $64.2$ & $81.8$ & $63.6$ & $66.5$ & $54.3$ & $62.1$  & $24$  \\

\cline{2-23} 
& \textbf{Lite-HDSeg} [Ours] 
& $\textbf{63.8}$ & $\textbf{92.3}$ & $40.0$ & $\textbf{55.4}$ & $37.7$ & $\textbf{39.6}$ & $59.2$ & $\textbf{71.6}$ & $\textbf{54.1}$ & $\textbf{93.0}$ & $\textbf{68.2}$ & $\textbf{78.3}$ & $\textbf{29.3}$ & $\textbf{91.5}$ & $\textbf{65.0}$ & $78.2$ & $\textbf{65.8}$ & $65.1$ & $\textbf{59.5}$ & $\textbf{67.7}$ & $20$  \rule{0pt}{2.3ex}\\

\hline
\end{tabular} 
}
\caption[CPNET]{IoU results on the SemanticKITTI test dataset for projection-based and point-wise methods. FPS measurements were taken using a single GTX 2080Ti GPU, or approximated if a runtime comparison was made on another GPU.}
\label{bigtable}
\vspace{-10px}
\end{table*}

\subsection{Experimental setup} 
We trained our network for $100$ epochs using stochastic gradient descent with the initial learning rate of $0.01$ which was decayed by $0.01$ after each epoch. The batch size was set to $6$ and the dropout probability to $0.2$. 
 The number of channels [$C_0 \rightarrow C_{13}$] 
 used in Lite-HDSeg architecture were assigned as $5$, $24$, $32$, $48$, $96$, $192$, $320$, $720$, $1280$, $1744$, $842$, $448$, $192$, $64$, respectively. The height and width of the projected image were set to $H = 64$, $W = 2048$. 
The values of $1.0$, $1.5$, $1.0$, $1.0$ were used for the weights of the total loss, $\alpha$, $\beta$, $\gamma$, $\lambda$, respectively. In the post-processing stage, the window size of the neighbor search was set to $5$, with $k=5$, $\sigma=1$, and a cutoff of $1m$.
Moreover, we applied augmentation, such as random rotation, transformation, flipping around the $y$-axis with probability of 0.5, to the data to avoid overfitting. 

\subsection{Results}
\begin{figure}[!htbp]
    \centering
    \includegraphics[width=0.49\textwidth]{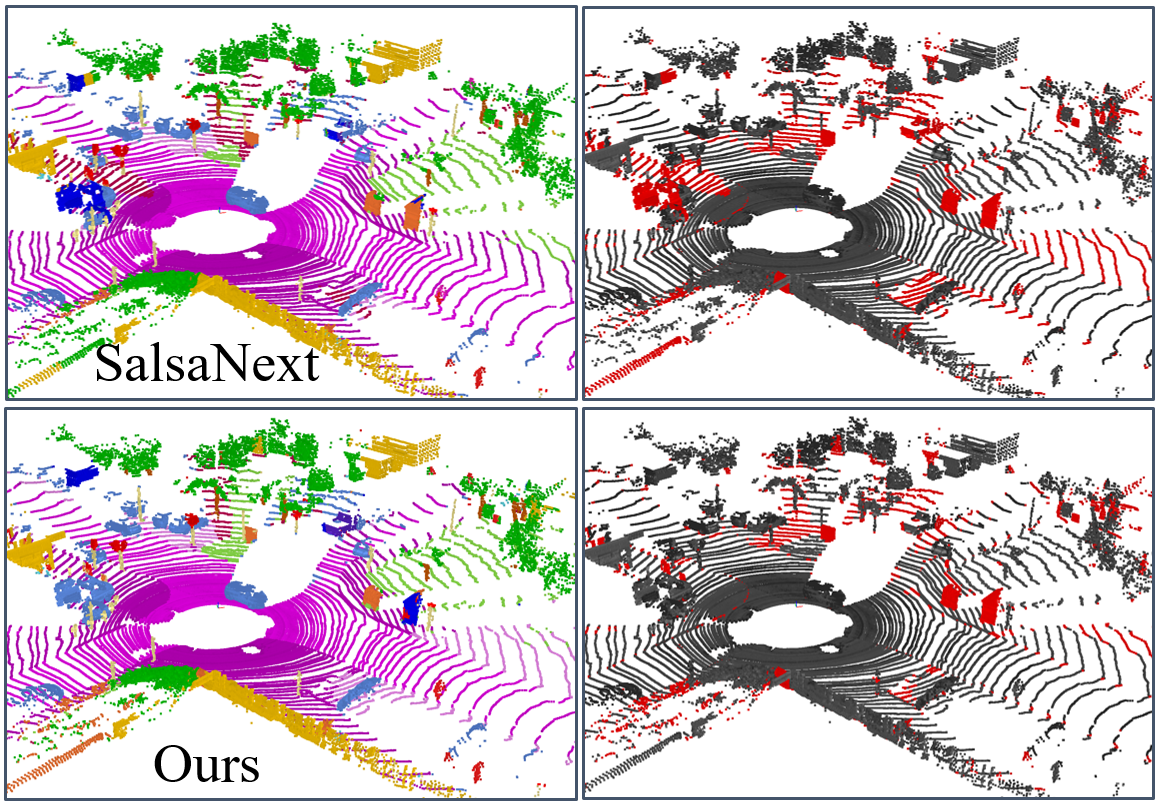}
    \caption{Comparison of prediction (left) and error map (right) of the proposed method vs. SalsaNext on SemanticKITTI validation set (sequence 08). 
	Color codes are: \crule[road]{0.2cm}{0.2cm} road \textbar
	\crule[side-walk]{0.2cm}{0.2cm} side-walk \textbar 
	\crule[parking]{0.2cm}{0.2cm} parking \textbar
	\crule[car]{0.2cm}{0.2cm} car \textbar 
	\crule[bicyclist]{0.2cm}{0.2cm} bicyclist \textbar
	\crule[pole]{0.2cm}{0.2cm} pole \textbar
	\crule[vegetation]{0.2cm}{0.2cm} vegetation \textbar
	\crule[terrain]{0.2cm}{0.2cm} terrain \textbar
	\crule[trunk]{0.2cm}{0.2cm} trunk \textbar
	\crule[building]{0.2cm}{0.2cm} building \textbar
	\crule[other-structure]{0.2cm}{0.2cm} other-structure \textbar
	\crule[other-object]{0.2cm}{0.2cm} other-object.}
    \label{fig:case-1}
    \vspace{-15px}
\end{figure}

The numerical comparison of the described techniques is illustrated in Table \ref{bigtable} for recent available and published methods of projection-based and point-wise categories. For each class, the best achieved accuracy in terms of \textbf{IoU} is highlighted among different methods. As shown in Table~\ref{bigtable}, the proposed method achieves the state-of-the-art performance on SemanticKITTI test set. Lite-HDSeg outperforms existing methods by $4.3$\% (mIoU) and most of the classes in terms of \textbf{IoU} within its category, while being real-time. It is worth noting that Lite-HDSeg not only outperforms all projection-based methods, but also outperforms real-time implementations of SPVNAS \cite{tang2020searching} with $63.7$ and $60.3$ mIoU accuracy (see Fig.~\ref{Iou_vs_Speed}). To better visualize the improvements made by our proposed model, a sample of qualitative result is shown in  Fig.~\ref{fig:case-1}. We compare SalsaNext vs. Lite-HDSeg in terms of the generated error map in the same data frame in Fig.~\ref{fig:case-1}, with Lite-HDSeg showing a noticeable improvement against SalsaNext.

\subsection{Ablation Study}
\label{sec:ablation}
In order to understand the effectiveness of different proposed techniques, a thorough ablation study has been done and the results are shown in Table~\ref{tab:Ablation} for Lite-HDSeg. The $\mathbf{mIoU}$ scores of all ablated networks are based on our training and the reported numbers are the $\mathbf{mIoU}$  for sequence $8$ on SemanticKITTI for all the experiments.

The ablation study starts with the baseline model, Salsanext \cite{cortinhal2020salsanext}, which is the closest method proposed in terms of performance to Lite-HDSeg. Next, we exchange the backbone with the proposed encoder-decoder in Section \ref{sec:proposed_method}. Afterwards, we add the ICM, CAM, boundary loss and MCSPN one by one, to show the effectiveness of each block. As it is shown in Table~\ref{tab:Ablation}, all the contributions stated in the table effectively improve our proposed encoder-decoder with the biggest jump of $2.3\%$ when adding MCSPN. It is worth noting that, although extra modules are proposed in the design of Lite-HDSeg, because of the novel reduction scheme in the complexity of the harmonic dense block (lite HD block), the overall computational complexity of Lite-HDSeg remains at a competitive range.

\begin{table}[h]
\begin{center}
\scalebox{0.8}{
\begin{tabular}{|c|cccccc|c | }
\hline \hline 
\multicolumn{1}{|c|}{\textbf{Architecture}} &
\multicolumn{1}{c}{\textbf{Backbone}} & 
\multicolumn{1}{c}{\textbf{ICM}} &
\multicolumn{1}{c}{\textbf{CAM}} &
\multicolumn{1}{c}{\textbf{\begin{tabular}[c]{@{}c@{}}Boundary\\ Loss\end{tabular}}}  & 
\multicolumn{1}{c|}{\textbf{MCSPN}} &
\multicolumn{1}{ c|}{\textbf{mIoU}} \\

 \hline \hline 
\multirow{1}{*}{Baseline}
 & SalsaNext & \xmark & \xmark & \xmark & \multicolumn{1}{c|}{\xmark}  & 56.9  \rule{0pt}{3ex}\\ \hline
 
\multirow{5}{*}{Lite-HDSeg}

 & Ours & \xmark & \xmark & \xmark & \multicolumn{1}{c|}{\xmark}  & 59.1  \rule{0pt}{3ex}\\ \cline{2-7}  

 & Ours &\checkmark  & \xmark & \xmark & \multicolumn{1}{c|}{\xmark}  & 59.4  \rule{0pt}{3ex}\\ \cline{2-7}  

 & Ours &\checkmark  & \checkmark  & \xmark & \multicolumn{1}{c|}{\xmark}  & 60.3  \rule{0pt}{3ex}\\ \cline{2-7}  

 & Ours &\checkmark  & \checkmark  & \checkmark & \multicolumn{1}{c|}{\xmark}  & 62.1  \rule{0pt}{3ex}\\ \cline{2-7}  

 & Ours &\checkmark  & \checkmark  & \checkmark & \multicolumn{1}{c|}{\checkmark}  & 64.4  \rule{0pt}{3ex}\\ \hline

\end{tabular}}
\end{center}
\caption{Ablative Analysis evaluated on SemanticKITTI dataset validation (seq 08). Our proposed backbone is the encoder-decoder introduced in Sections \ref{encoder} and \ref{decoder}.}
\label{tab:Ablation}
\end{table}

\section{Conclusions}
\label{sec:conclusion}

In this paper, we presented Lite-HDSeg, a novel real-time CNN model for semantic segmentation of a full $3$D LiDAR point cloud. Lite-HDSeg has a new encoder-decoder architecture based on light-weight harmonic dense convolutions and residual blocks with a novel contextual module called ICM and Multi-class SPN. We trained our network with boundary loss to emphasize the semantic boundaries. To show the performance of the proposed method, Lite-HDSeg was evaluated on the public benchmark, SemanticKITTI. Experiments show that the proposed method outperforms all \textit{real-time} state-of-the-art semantic segmentation approaches in terms of accuracy ($\mathbf{mIoU}$), while being less complex. More specifically, Lite-HDSeg introduces a novel design with accuracy improvements from each of the blocks, i.e., the new context module (ICM), the new encoder-decoder, MCSPN and boundary loss, to benefit not only semantic segmentation task, but also instance segmentation and object detection.

\bibliographystyle{IEEEtran}
\bibliography{summarybib}

\end{document}